\documentclass[10pt,twocolumn,letterpaper]{article}

\usepackage{iccv}
\usepackage{times}
\usepackage{epsfig}
\usepackage{graphicx}
\usepackage{amsmath}
\usepackage{amssymb}

\usepackage{booktabs}

\usepackage{mathtools}
\usepackage{amsthm}

\usepackage[utf8]{inputenc} %
\usepackage[T1]{fontenc}    %
\usepackage{url}            %
\usepackage{booktabs}       %
\usepackage{amsfonts}       %
\usepackage{nicefrac}       %
\usepackage{microtype}      %
\usepackage{xcolor}         %
\usepackage{bm}
\usepackage[textsize=tiny]{todonotes}

\usepackage{threeparttable}
\usepackage{makecell}
\usepackage{paralist}
\usepackage{multirow}

\usepackage[pagebackref=true,breaklinks=true,letterpaper=true,colorlinks,bookmarks=false]{hyperref}

\iccvfinalcopy %

\def\iccvPaperID{6861} %

\ificcvfinal\fi

\usepackage[capitalize]{cleveref}
\crefname{section}{Sec.}{Secs.}
\Crefname{section}{Section}{Sections}
\Crefname{table}{Table}{Tables}
\crefname{table}{Tab.}{Tabs.}

\begin{document}

\title{Adaptive Calibrator Ensemble for Model Calibration under Distribution Shift}

\author{
Yuli Zou$^{1}$ \quad Weijian Deng$^{2}$ \quad Liang Zheng$^{2}$\\
$^{1}$The Hong Kong Polytechnic University \quad $^{2}$The Australian National University\\
}

\maketitle
\ificcvfinal\thispagestyle{empty}\fi

\begin{abstract}
Model calibration usually requires optimizing some parameters ({e.g.}, temperature) {w.r.t} an objective function ({e.g.}, negative log-likelihood).
In this paper, we report a plain fact that the objective function is influenced by calibration set difficulty, {i.e.}, the ratio of the number of incorrectly classified samples to that of correctly classified samples \footnote{To possibly facilitate reader understanding, we point out that the difficulty of a dataset (with respect to a classifier) shares the same meaning of classifier accuracy on this dataset. We define ``difficulty'' to describe the property of a dataset (\emph{i.e.,} its OOD degree), instead of using ``accuracy'' which describes the performance of the classifier on a dataset.}.
If a test set has a drastically different difficulty level from the calibration set, a phenomenon out-of-distribution (OOD) data often exhibit: the optimal calibration parameters of the two datasets would be different, rendering an {optimal} calibrator on the calibration set {suboptimal} on the OOD test set and thus degraded calibration performance.   %
With this knowledge, we propose a simple and effective method named adaptive calibrator ensemble (ACE) to calibrate OOD datasets whose difficulty is usually higher than the calibration set. Specifically, two calibration functions are trained, one for in-distribution data (low difficulty), and the other for severely OOD data (high difficulty). To achieve desirable calibration on a new OOD dataset, ACE uses an adaptive weighting method that strikes a balance between the two extreme functions. When plugged in, ACE generally improves the performance of a few state-of-the-art calibration schemes on a series of OOD benchmarks. Importantly, such improvement does {not} come at the cost of the in-distribution calibration performance.
\end{abstract}

\section{Introduction}
\label{sec:intro}
Model calibration aims to connect the neural network output with uncertainty. A common practice is to find optimal parameters against certain objective functions on a held-out calibration set, to obtain an optimized \emph{calibrator}. 
In this paper, we focus on post-hoc calibration methods, which require training a calibration mapping function to rescale the confidence scores of a trained neural network to make it calibrated \cite{guo2017calibration,gupta2021calibration,kull2019beyond}. A popular technique is Temperature Scaling \cite{guo2017calibration}, which optimizes model temperature by minimizing the negative log-likelihood (NLL) loss. 

Post-hoc calibration methods generally work well when calibrating in-distribution test sets. However, oftentimes their calibration performance drops significantly when being tested on an out-of-distribution (OOD) test set \cite{ovadia2019can}. For example, temperature scaling has shown to be ineffective under distribution shift in some scenarios \cite{ovadia2019can}. This problem happens because the test environment (OOD) is different from the training environment due to factors like sample bias and non-stationarity. This paper thus aims to improve post-hoc calibration methods by producing reliable and predictive uncertainty under distribution shifts.

In the community, there exist a few works studying the OOD calibration problem \cite{salvador2021improved,tomani2021post,wang2020transferable}.
They typically aim to make amendments to the calibration set to let it approximate the OOD data in certain aspects \cite{salvador2021improved,tomani2021post}. Nevertheless, these techniques are typically not adaptive to the test dataset, that is, the calibration set transformation process cannot automatically adjust to the test set. In our experiment, we observe that they improve calibration on some OOD datasets but significantly lead to decreased in-distribution calibration performance.
In this regard, while TransCal \cite{wang2020transferable} can perform domain adaptation according to the test domain, it needs to be re-trained for every new test set.

In this paper, our contributions are mainly in two aspects.
\textbf{First}, we provide a new perspective to understand calibration failure on out-of-distribution datasets.
Specifically, we show that \textbf{the calibration objective is dependent on the dataset difficulty.}
When the calibration set have the same distribution with the test set, it has low difficulty, and thus the calibrator learned on the calibration set would be effective on the test set \cite{guo2017calibration,gupta2021calibration,kull2019beyond}. 
However, out-of-distribution test sets usually exhibit a different (in fact, higher) difficulty level compared with the calibration set because of the distribution gap. 
Under this case, the optimal calibration functions are different between the calibration set and OOD test sets. 
That is, a calibrator that optimized on the calibration set would not be optimal on OOD data and thus it would achieve poor calibration performance. 

\textbf{Second}, to achieve robust calibration under distribution shifts, we propose a simple but effective method named adaptive calibrator ensemble (ACE).
It adaptively integrates two predefined calibrators: 1) one trained on an easy in-distribution dataset, and 2) the other trained on a severely OOD data set with high difficulty. By estimating how much a new test set deviates from the high-difficulty calibration set, we compute a test adaptive weight to balance the force between the two calibrators.
We show that our proposed ACE method improves three existing post-hoc calibration algorithms such as Spline \cite{gupta2021calibration} on commonly used OOD benchmarks. Moreover, our method does \emph{not} have compromised calibration performance for in-distribution data.

\section{Related Work}
\label{sec:related}
\textbf{Post-hoc calibration} calibrates a trained neural network by rescaling confidence scores \cite{allikivi2019non,guo2017calibration,gupta2021calibration,joy2022sample,kull2017beta,kull2019beyond,liu2020simple,naeini2016binary,nixon2019measuring,rahimi2020intra,tian2021geometric,van2020uncertainty,wenger2020non,zadrozny2001obtaining,zadrozny2002transforming}.
For example, as a multi-class extension of Platt scaling, vector scaling and matrix scaling \cite{guo2017calibration} introduce a linear layer to transform the logits vector to calibrate the network outputs.
Spline \cite{gupta2021calibration} obtain a recalibration function via spline-fitting, which directly maps the classifier outputs to the calibrated probabilities.
Dirichlet \cite{kull2019beyond} propose a multi-class calibration method, derived from Dirichlet distributions.
Rahimi \etal \cite{rahimi2020intra} propose a general post-hoc calibration function that can preserve the top-$k$ predictions of any deep network via intra order-preserving function. 
Our work seeks to improve the OOD performance of existing post-hoc calibrators such as vector scaling, temperature scaling, and spline, through an ensemble mechanism.

\textbf{Out-of-distribution calibration.} A few works study calibration under distribution shift \cite{salvador2021improved,wang2020transferable}.
To improve the post-hoc calibration under distribution shift, some researches \cite{salvador2021improved,tomani2021post} propose to modify the calibration set to represent a generic distribution shift.
Moreover, prediction uncertainty is studied in  \cite{krishnan2020improving}. Based on the uncertainty, an “accuracy versus uncertainty” calibration loss is proposed to encourage a model to be certain on correctly classified samples and uncertain on inaccurate samples. In comparison, our method is based on \emph{whether samples are correctly or incorrectly classified (i.e., difficulty) rather than uncertainty}. We find difficulty is an important factor for OOD calibration failure.
Furthermore, TransCal \cite{wang2020transferable} uses unsupervised domain adaptation to improve temperature scaling. This method has a high computational cost because, 1) it needs an additional domain adaptation training process, and 2) every time it meets a new test set, the domain adaptation model needs to be re-trained.
{
Gong \etal \cite{gong2021confidence} study the calibration under domain generalization setting where they develop calibration methods on calibration sets from \textit{multiple} domains.
}
We contribute from a different perspective to the existing literature. We provide insight into the role of dataset difficulty on the failure of existing algorithms on OOD data.
We then propose a simple and effective ensemble strategy to improve post-hoc calibrators in a test set adaptive manner.

\section{Methodology}
\label{sec:method}
\subsection{Preliminaries}
\textbf{Neural network notations.} {Considering the task of calibrating neural networks for $n$-way classification, let us define $[n] \coloneqq \{1,\dots,n\}$, $\mathcal{X}\subseteq \mathbb{R}^d$ be the domain, $\mathcal{Y}  = [n]$ be the label space, and $\Delta_n$ denote the $n-1$ dimensional unit simplex.
Given a training dataset $\mathcal{D}_{tr}$ of independent and identically distributed (i.i.d.) samples drawn from an unknown distribution $\pi$ on $\mathcal{X} \times \mathcal{Y}$, we learn a probabilistic predictor $\mathbf{\phi}:\mathbb{R}^d \to \Delta_n$.
We assume that $\mathbf{\phi}$ can be expressed as the composition $\mathbf{\phi} \eqqcolon \mathbf{sm} \circ \mathbf{g}$, with $\mathbf{g}: \mathbb{R}^d \to \mathbb{R}^n$ being a non-probabilistic $n$-way classifier and $\mathbf{sm}: \mathbb{R}^n \to \Delta_n$ being the softmax operator
$\mathbf{sm}_i(\mathbf{z}) = \frac{\exp(\mathbf{z}_i)}{\sum_{j=1}^n \exp(\mathbf{z}_j)}$, for $i \in \mathcal{Y}$, 
where the subscript~$_i$ denotes the $i$-th element of a vector.
We say $\mathbf{g}(\mathbf{x})$ is the \emph{logits} of $\mathbf{x}$ with respect to $\mathbf{\phi}$. }

\textbf{Definition of a calibrated network.} 
When queried at $(\mathbf{x}, y)\in \mathcal{X} \times \mathcal{Y}$ sampled from an unknown distribution $\pi$, the probabilistic predictor $\mathbf{\phi}$ returns $\hat{y} \eqqcolon \arg\max_{i}$ $\mathbf{\phi}_{i}(\mathbf{x})$ as the predicted label and $\hat{p} \eqqcolon \max_i \mathbf{\phi}_{i} (\mathbf{x})$ as the associated confidence score.  
{We say $\mathbf{\phi}$ is \emph{perfectly calibrated} with respect to $\pi$, if $\hat{p}$ is expected to represent the true probability of correctness.}
Formally, a perfectly calibrated model satisfies $\mathbb{P}(\hat{y}=y|\hat{p}=p)=p$ for any $p \in [0,1]$.
In practice, we commonly use the Expected Calibration Error (ECE)  \cite{guo2017calibration}  as the calibration performance metric. It  first groups all samples into $M$ equally interval bins $\{B_m\}_{m=1}^M$ with respect to their confidence scores, and then calculates the expected difference
between the accuracy and average confidence: $\mathrm{ECE} =\sum_{m=1}^M\frac{|B_m|}{n}|\text{acc}(B_m)-\text{avgConf}(B_m)|$, where $n$ denotes the number of samples. 

{\textbf{Post-hoc calibration} learns a post-hoc calibration function $\mathbf{f}: \mathcal{R}^n\to\mathcal{R}^n$ such that the new probabilistic predictor $\mathbf{\phi}_c \coloneqq \mathbf{sm} \circ \mathbf{f} \circ \mathbf{g}$ is better calibrated \emph{and} tries to keep a similar (or same) accuracy of the original network $\mathbf{\phi}$}.

\subsection{Post-hoc Calibration Function Is Influenced by Calibration Set Difficulty}
\label{sec:d_influence}

\textbf{Post-hoc calibration loss function.} 
Assume we have a held-out calibration dataset $\mathcal{D}_c= \{(\mathbf{x}^{i}, y^{i})\}_{i=1}^N$ with i.i.d samples from the unknown distribution $\pi$ on $\mathcal{X} \times \mathcal{Y}$ and a calibration function $\mathbf{f}$ parameterized by some vector $\mathbf{\theta}$. The empirical calibration loss is generally defined as,
\begin{equation}
\label{cal_loss}
\frac{1}{N} \sum_{i=1}^N \ell(y^{i}, \mathbf{f}(\mathbf{z}^{i})) + \frac{\lambda}{2} ||\mathbf{\theta}||^2, 
\end{equation}

where $\mathbf{z}^i = \mathbf{g}(\mathbf{x}^i)$, 
$\ell:\mathcal{Y}\times\mathcal{R}^n\to\mathcal{R}$ is a cost function, and $\lambda \geq 0$ is the regularization weight. $\ell(\cdot, \cdot)$ is the network classification loss. Following existing literature, we employ the commonly used negative log-likelihood (NLL) loss:
\begin{equation}
\label{nll}
    \ell(y, \mathbf{f}(\mathbf{z})) = -\log(\mathbf{sm}_y(\mathbf{f}(\mathbf{g}(\mathbf{x})))),
\end{equation}
where $\mathbf{sm}$ is softmax operator, and $\mathbf{sm}_y$ is its $y$-th element.

\begin{figure}
    \centering
    \includegraphics[width=0.85\linewidth]{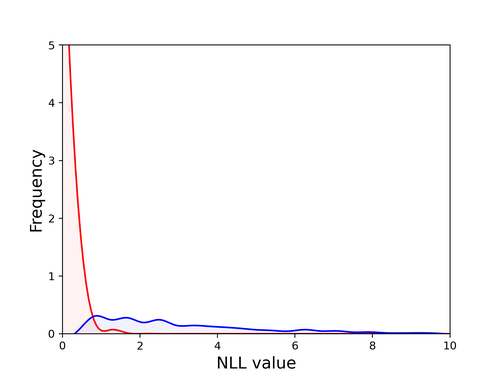}
    \caption{
    \textbf{NLL values of correctly and incorrectly classified samples.} We use ResNet-$152$ on the in-distribution ImageNet calibration set (described in Section \ref{setup}) and plot NLL probability density of the two types of samples. We clearly observe that correctly classified samples generally have a much lower NLL value.
    }
    \label{fig:nll}
    \vskip -0.15in
\end{figure}

\textbf{Plain fact: individual samples matter in the classification loss.} %
Apparently, a major component in the calibration objective (Eq. \ref{cal_loss}) is the model classification loss (\emph{e.g.,} the commonly used NLL loss, Eq. \ref{nll}). If a sample is correctly classified, %
the classification loss will likely return a small value; If a sample is incorrectly classified, %
there will likely be a high loss value. Therefore, \emph{whether an individual sample is correctly classified or not would lead to quite different classification loss values}. 

We conduct an empirical analysis to verify this conclusion. %
Specifically, we use ResNet-152 trained on  ImageNet \cite{deng2009imagenet}. The NLL values of these samples are computed on the calibration set (described in Section \ref{setup}), and summarily drawn in Fig. \ref{fig:nll}. It is clearly shown that the NLL values of correctly classified samples are close to $0$ while those of incorrectly classified samples are significantly greater.

\textbf{Collectively, calibration set difficulty influences calibration optimization.}
To illustrate this point, we use NLL as an example, which is a commonly used classification loss.
Given that the two types of samples have different NLL values, we decompose the NLL loss into two parts: 
\begin{equation}
    \ell_T(y, \mathbf{f}(\mathbf{z}))=-\frac{1}{N_T}\sum_i^{N_T}\log(\mathbf{sm}_{y^i}(\mathbf{f}(\mathbf{g}(\mathbf{x}^i)))),
    \label{eq:LT}
\end{equation}
where $\arg\max \mathbf{sm}(\mathbf{g}(\mathbf{x}^i))= y^i$, and,
\begin{equation}
    \ell_F(y, \mathbf{f}(\mathbf{z}))=-\frac{1}{N_F}\sum_i^{N_F}\log(\mathbf{sm}_{y^i}(\mathbf{f}(\mathbf{g}(\mathbf{x}^i)))),
        \label{eq:LP}
\end{equation}
where $\arg\max \mathbf{sm}(\mathbf{g}(\mathbf{x}^i))\neq y^i$.
In Eq. \ref{eq:LT} and Eq. \ref{eq:LP}, $N_T$ and $N_F$ note the numbers of correctly and incorrectly classified samples, respectively. %
By adjusting $\frac{N_F}{N_T}$,  %
the overall NLL value changes, which will affect the optimized calibration parameters $\bm{\theta}$ ({\emph{a.k.a.} the calibration function}). 

Formally, we define the \emph{difficulty} of a dataset as $\frac{N_F}{N_T}$. 
Note that, the difficulty of a dataset (with respect to a classifier) shares the same meaning as classifier accuracy on this dataset.
The above analysis indicates that \emph{optimized calibration parameters are affected by the difficulty  of the calibration set}: \textbf{1)} $\mathbf{\theta}$ trained on a more difficult calibration set tends to have a larger classification loss values (Eq. \ref{nll}) and thus a larger calibration loss (Eq. \ref{cal_loss}). \textbf{2)} $\mathbf{\theta}$ trained on an easier calibration set likely corresponds to a smaller classification loss (Eq. \ref{nll}) and thus a lower calibration loss (Eq. \ref{cal_loss}).

\begin{figure}
    \centering
    \includegraphics[width=0.85\linewidth]{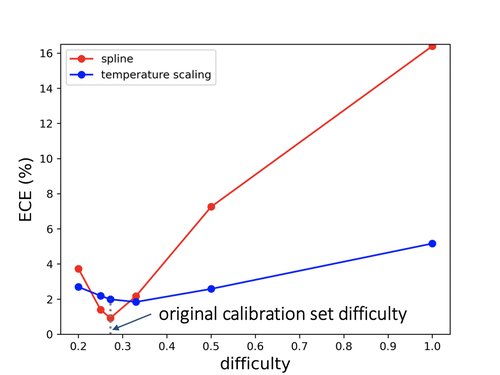}
    \caption{
     \textbf{The impact of calibration set difficulty ($\frac{N_F}{N_T}$) on calibration performance (ECE).} We manually select images from the in-distribution calibration set to create new calibration sets of various difficulty levels. The difficulty of the original calibration set is $0.2723$. We use the ResNet-$152$ model and an in-distribution test set from ImageNet. We evaluate two methods (temperature scaling, and Spline), and at the same time, mark the difficulty of the original in-distribution calibration set (gray vertical dotted line).
     We find the calibration sets having similar difficulty to the original will lead to good calibration performance and vice versa.
    }
    \label{fig:diff}
    \vskip -0.15in
\end{figure}

We empirically verify the above conclusion in Fig. \ref{fig:diff}%
, where we create calibration sets with various levels of difficulty ($\frac{N_F}{N_T}$) and mark the difficulty level of the original calibration set. It indicates that 
calibration set difficulty indeed influences ECE of two calibration methods: temperature scaling (NLL) \cite{guo2017calibration}, Spline (KS-error) \cite{gupta2021calibration}. Moreover, when testing the original in-distribution data,
if the difficulty of the created calibration dataset is similar,  %
the two calibration methods generally have good calibration performance. 
However, calibration performance is poorer when the \emph{difficulty} of created calibration dataset is very different from that of the original calibration dataset.

The above analysis mostly uses the NLL loss as an example,
but can also apply to some other classification loss functions (\emph{e.g.}, the KS-error used in Spline is verified in Fig. \ref{fig:diff}, the cross-entropy loss and the focal loss).
These loss functions are usually influenced by individual samples, and thus collectively the dataset difficulty would eventually impact the calibration performance.

\subsection{Calibration Set Difficulty Influences Out-of-distribution Calibration}

Having analyzed that calibration set difficulty influences the calibration performance on in-distribution test sets, we provide a tentative explanation of \textit{why calibrators trained on in-distribution data fail on OOD test sets}. 
Essentially, an OOD test set usually has a different difficulty level from the in-distribution calibration set. In fact, the OOD difficulty level is usually higher, \emph{i.e.,} there is a higher percentage of incorrectly classified samples, because of the domain gap problem \cite{ben2010theory,deng2021labels,mansour2009domain}. Therefore, the calibration mapping function $\mathbf{f}$ that an OOD test set needs is different from the in-distribution calibration set. When training a calibrator on in-distribution data, its performance would thus be suboptimal on the OOD test data. 

Moreover, our reasoning also helps understand why some existing OOD calibration methods have compromised calibration performance on in-distribution test sets. Specifically, these methods, \emph{e.g.,} Perturbation \cite{tomani2021post}, obtain their mapping functions on some modified in-distribution calibration set (\emph{e.g.,} adding Gaussian noise), which to some extent mimics the OOD test set. However, this modification operation is not adaptive, that is, they do not change \emph{w.r.t} the test set. When the test set changes to an in-distribution one, its optimal calibration parameters would be different from those obtained from the modified calibration set. This is possible because of different difficulty levels. 

\subsection{Adaptive Calibrator Ensemble}\label{method:ensemble}

\textbf{Overview.} To achieve desirable calibration under distribution shifts, we propose a simple and effective method called Adaptive Calibrator Ensemble (ACE).
Using an in-distribution calibration set as input, ACE outputs an OOD calibrator as if having been trained on a calibration set with a proper difficulty level. %
To do so, we first seek two calibration sets with extreme difficulty levels: an in-distribution difficulty level (easy) and a high difficulty level (hard). We then use an adaptive weighting scheme to fuse the output of calibrators trained on the two extreme calibration sets. 

\textbf{Finding two datasets with extreme difficulty levels.} Straightforwardly, we secure the ``easy'' one as the in-distribution calibration set itself $\mathcal{D}_o$. To obtain the ``hard'' calibration set $\mathcal{D}_h$, we perform sampling on $\mathcal{D}_o$ aiming to increase the difficulty. 
Specifically, we apply the classifier on the in-distribution calibration set to find correctly classified samples, incorrectly classified samples, and their numbers $N_{T}^o$ and $N_{F}^o$ ($N_{T}^o$ is usually greater than $N_{F}^o$). To create $\mathcal{D}_h$, we calculate its $N_T$ and $N_F$ as follows, $N_F^h = N_{F}^o, N_T^h = N_{F}^o/d$, where $d \in (0, \infty)$ is a pre-defined difficulty level (hyperparameter). We then randomly sample $\mathcal{D}_o$ to achieve this difficulty level.
When $d$ is relatively large\footnote{By default, we set $d=10$, which means $10$ times more incorrectly classified samples than correct ones. Notice that we set $d=9$ for CIFAR-$10$-C, which equals the randomly classified result.}, the calibration set contains many more incorrectly classified samples than correctly classified ones, allowing us to have the desired calibration set $\mathcal{D}_h$, which is considered seriously out-of-distribution and hard.

\textbf{Training two calibrators on the two extreme datasets.} On \emph{each} of the obtained the easy and the hard calibration sets $\mathcal{D}_o$ and $\mathcal{D}_h$, we train a calibrator. Let $\mathbf{g}$ denote the deep learning model.
For calibration dataset $\mathcal{D}_o= \{(\mathbf{x}^{i}, y^{i})\}_{i=1}^{N_{o}}$, where $N_{o}$ means the number of samples of $\mathcal{D}_o$, we train a calibration function $\mathbf{f}_o$, and the calibrated logits are denoted as $\mathbf{z}_{o}^i=\mathbf{f}_o(\mathbf{z}_\text{ori}^i)$. 
Here, $\mathbf{z}_\text{ori}^i$ is the original uncalibrated logits for a new test set that is either in-distribution or OOD.
Similarly, for calibration set $\mathcal{D}_h= \{(\mathbf{x}^{i},y^{i})\}_{i=1}^{N_{h}}$, where $N_{h}$ means the number of samples of $\mathcal{D}_h$, we train a calibration function $\mathbf{f}_h$, the calibrated logits is $\mathbf{z}_{h}^i=\mathbf{f}_h(\mathbf{z}_\text{ori}^i)$.

\textbf{An adaptive method to ensemble outputs of the two calibrators.} Given $\mathcal{D}_o$ and $\mathcal{D}_h$, we intuitively speculate that the difficulty of a usual out-of-distribution test set would be positioned in between. As such, we propose to compute an adaptive weight $\alpha$ to balance the difficulty of these two outputs produced by calibrators, then the final output $\mathbf{z}_\text{cal}$ is:
\begin{equation}
\mathbf{z}_\text{cal} = \alpha\cdot\mathbf{z}_o + (1-\alpha)\cdot\mathbf{z}_h.
\label{eq:weighted_sum}
\end{equation}

In designing a reasonable weight $\alpha$, we request it to be test-set-adaptive. \textbf{First}, when the distribution of an OOD test set is similar to the original calibration set $\mathcal{D}_o$, $\alpha \rightarrow 1$, so that the system reduces to in-distribution calibrator $\mathbf{z}_o$;
\textbf{Second}, when a test set is seriously out-of-distribution, $\alpha \rightarrow 0$.

Moreover, \cite{guillory2021predicting} suggest that the average confidence score could serve as an unsupervised indicator of the degree of how out-of-distribution a test set is. So given an unlabeled test set $\mathcal{D}_\text{test}$, we can estimate an approximate OOD degree of this test set. Here, we compute the ad-hoc weight $\alpha$ as,
\begin{equation}
\label{eq:weight}
    \alpha = \frac{\text{avgConf}(\mathcal{D}_\text{test})}{ \text{avgConf}(\mathcal{D}_{o})},
\end{equation}
where $\text{avgConf}(\cdot)$ calculates the average confidence score of a dataset. In the experiment, we will evaluate some fixed values of $\alpha$, which are useful on some occasions but less so on others. Moreover, being fixed implies that it does not work for in-distribution data unless it is fixed to $1$.

\textbf{The ensemble scheme works efficiently.}
To illustrate how ACE ensembles the two calibrators, here we use Temperature Scaling \cite{guo2017calibration} as an example whose calibration function is $\mathbf{f}(\mathbf{z})=\mathbf{T}\cdot\mathbf{z}$ where $\mathbf{T}$ is a learnable scalar parameter.
Let $\mathbf{T_o}$ and $\mathbf{T_h}$ denote the temperature value which learned on the easy calibration set $\mathcal{D}_o$ and the hard calibration set $\mathcal{D}_h$, respectively. $\mathbf{z}_\text{ori}$ is the uncalibrated logits of test set. Referring to Eq. \ref{eq:weighted_sum}, the calibrated logits of test set $\mathbf{z}_\text{cal}$ is:
\begin{equation}
\begin{split}
        \mathbf{z}_\text{cal} 
        &= \alpha \cdot {\mathbf{z}_\text{ori}}\cdot{\mathbf{T}_o} + (1-\alpha) \cdot {\mathbf{z}_\text{ori}}\cdot{\mathbf{T}_h} \\
        &= \mathbf{z}_\text{ori}\cdot({\alpha}\cdot{\mathbf{T}_o} + ({1-\alpha})\cdot{\mathbf{T}_h}).
\end{split}
\end{equation}
Thus the equivalent value of temperature $\mathbf{T}_\text{cal}$ which has $\mathbf{z}_\text{cal} = {\mathbf{z}_\text{ori}}\cdot{\mathbf{T}_\text{cal}}$ can be computed as:
\begin{equation}
\label{eq:t_cal}
\begin{split}
        \mathbf{T}_\text{cal} = {\alpha}\cdot{\mathbf{T}_o} + (1-\alpha)\cdot{\mathbf{T}_h}.
\end{split}
\end{equation}
According to Eq. \ref{eq:t_cal}, we show that the output-space ensemble (Eq. \ref{eq:weighted_sum}) is equal to the weight-space ensemble of two calibrators. 
Moreover, weight-space ensemble methods have shown superior performance and robustness gains over single models \cite{havasi2021training, lakshminarayanan2017simple, premchandar2022unified, wen2020batchensemble, zhang2020mix}.
Therefore, the outputs produced by our ensemble scheme shows to have better calibration performance than single calibrator produces.

\subsection{Discussion}
\label{discussion}
\textbf{Difficulty is a relative concept.} 
Despite being formulated as $\frac{N_F}{N_T}$, difficulty also depends on the model or classifier. For stronger models, the difficulty level would be lower (even $N_F = 0$) and vice versa. 
In this paper, we assume fixed models and choose not to put the model as a subscript in the definition of difficulty for simplicity.

\textbf{Domain gap \emph{vs.} difficulty.} 
Domain gap is used to describe the distribution difference between domains and certainly exists between an OOD test set and the calibration set. 
Therefore, a possible way to calibrate OOD data is to find a dataset with similar distribution to the OOD test set, which is essentially reflected in \cite{ salvador2021improved,tomani2021post}.
Our paper points out a new way to craft the domain gap by modifying the difficulty of the calibration set. In fact, domain gap is a complex phenomenon and related to many factors aside from difficulty, so it would be interesting to investigate other factors which can help OOD calibration.

\textbf{An alternative method.} %
We emphasize the main contribution is to report that calibration set difficulty is influential on OOD calibration performance. The designed method, in comparison, is more from an intuitive perspective. There might be other alternatives. For example, we could use the average confidence of a dataset (we use it in Eq. \ref{eq:weight} to calculate $\alpha$ instead) to estimate its difficulty and create a calibration set that has a closer difficulty level to the OOD test dataset. 
We show this alternative also gives improvement over some baselines.
(Please refer to the supplemental material for more details.)

\textbf{Potential limitation and further direction.} Our weighting method (Eq. \ref{eq:weighted_sum}) assumes that an OOD test set sits between $\mathcal{D}_o$ and $\mathcal{D}_h$ in terms of difficulty. This assumption should be valid for most cases in practice because the difficulty of $\mathcal{D}_o$ is very low and that of $\mathcal{D}_h$ is very high (we use $d=10$ by default, which translates to $9.1\%$ top-$1$ accuracy).  
We empirically observe that $d=10$ is effective, which translates to an accuracy of 9.09\%. We believe a dataset with 9.09\% accuracy is difficult enough to cover a wide range of test sets.
In addition, distribution shift occurs in a variety of ways \cite{hendrycks2021many,taori2020measuring}. 
There might exist scenarios (\emph{e.g.,} adversarial attack) where the confidence score is less effective in describing the distribution shift.
In such cases, our method might not be able to achieve significant improvement over existing algorithms. 
In fact, it would be interesting to explore other potential ways to characterize distribution discrepancy. 
{
Furthermore, in realistic application scenarios, we may have access to calibration datasets from multiple domains \cite{gong2021confidence}. To better use these data, one potential way is to learn a ACE model on each calibration set. Then, we ensemble the results of all learned ACE models for a given unknown test set. 
We evaluate our proposed ACE method under the domain generalization setting in the supplemental material.
}

\setlength{\tabcolsep}{10pt}
\begin{table*}[t]
\begin{small}
\caption{OOD calibration performance of our method (ACE) integrated with three post-hoc methods: vector scaling, temperature scaling (Temp. Scaling), and Spline. ECE ($25$ bins, \%) for top-$1$ predictions is reported. We use \textbf{{ResNet-$152$}} on various image classification datasets with \emph{various distribution shifts}. For each column, the lowest number is in \textbf{bold} and the second lowest \underline{underlined}.
Our method (ACE) effectively improves the post-hoc methods on $15$ out of $18$ occasions.
\textcolor{green}{$\blacktriangle$}/\textcolor{red}{$\blacktriangledown$} denotes %
ECE is lower / higher than the post-hoc method when being used alone, with statistical significance (p-value $\textless 0.05$) based on the two-sample t-test.
}
\begin{center}
\begin{tabular}{lcccccccc}
\toprule
 Methods &  ImgNet-V2-A &  ImgNet-V2-B &  ImgNet-V2-C &  ImgNet-S &  ImgNet-R &  ImgNet-Adv\\
\midrule
uncalibrated & $9.5016$ & $6.2311$ & $4.3117$ & $24.6332$  & $17.8621$ & $50.8544$ \\
\midrule
Vector Scaling &  $6.8068$ & $4.2184$ & $2.9258$ & $20.3726$  & $14.5037$ & $44.7593$\\
+ ACE & \makecell[c]{$5.6291$ \\$_{\pm0.0397}$ \textcolor{green}{$\blacktriangle$}}& \makecell[c]{$3.7742$ \\$_{\pm0.0237}$ \textcolor{green}{$\blacktriangle$}}& \makecell[c]{$3.1141$ \\$_{\pm0.0150}$ \textcolor{red}{$\blacktriangledown$}} & \makecell[c]{$15.8747$ \\ $_{\pm0.0252}$ \textcolor{green}{$\blacktriangle$}}& \makecell[c]{$10.6343$ \\ $_{\pm0.0356}$ \textcolor{green}{$\blacktriangle$}} & \makecell[c]{ $40.5773$ \\ $_{\pm0.0491}$ \textcolor{green}{$\blacktriangle$}}\\

\midrule
Temp. Scaling & $4.4413$ & $2.7309$ & $1.6831$ & $15.7879$  & $10.4797$ & $42.6302$ \\
+ ACE & \makecell[c]{$\underline{3.5615}$ \\ $_{\pm0.0028}$ \textcolor{green}{$\blacktriangle$}} & \makecell[c]{$2.5692$ \\ $_{\pm0.0013}$ \textcolor{green}{$\blacktriangle$}} & \makecell[c]{$1.7021$ \\ $ _{\pm0.0001}$ \textcolor{red}{$\blacktriangledown$}}& \makecell[c]{ $\underline{10.3915}$ \\$_{\pm0.0092}$ \textcolor{green}{$\blacktriangle$}} &\makecell[c] { $\underline{6.7458}$ \\ $_{\pm 0.0083}$ \textcolor{green}{$\blacktriangle$} }&\makecell[c]{  $\underline{38.0651}$\\$_{\pm0.0114}$ \textcolor{green}{$\blacktriangle$}} \\
\midrule
Spline& $4.5321$ & $\textbf{1.8034}$ & $\underline{1.3357}$ & $19.6392$  & $13.1116$ & $45.3623$\\
+ ACE & \makecell[c]{ $\textbf{2.8201}$ \\$_{\pm0.0283}$ \textcolor{green}{$\blacktriangle$}} & \makecell[c]{ $\underline{2.0235}$\\$_{\pm0.0154}$  \textcolor{red}{$\blacktriangledown$}} & \makecell[c]{ ${\textbf{1.0550}}$\\$_{\pm0.0092}$ \textcolor{green}{$\blacktriangle$}} &\makecell[c]{$\textbf{6.9264}$\\$_{\pm0.0864}$ \textcolor{green}{$\blacktriangle$}} &\makecell[c]{  $\textbf{6.8533}$\\$_{\pm0.0011}$ \textcolor{green}{$\blacktriangle$}} &\makecell[c] { $\textbf{31.0926}$\\$_{\pm0.0422}$ \textcolor{green}{$\blacktriangle$}}\\
\bottomrule
\end{tabular}
\label{resnet}
\end{center}
\end{small}
\vskip -0.2in
\end{table*}

\section{Experiment}
\label{exp}

\subsection{Experimental Setup}\label{setup}

\textbf{Neural Networks.} 
We consider both convolutional and non-convolutional networks. %
Specifically, we use ResNet-$152$ \cite{he2016deep}, {ViT-Small-Patch$32$-$224$ \cite{dosovitskiy2020image} and Deit-Small-Patch$16$-$224$ \cite{deit}}. The three networks are either trained or fine-tuned on the ImageNet training set \cite{deng2009imagenet}.

\textbf{Calibration set and in-distribution test set.}
Following the protocol in \cite{gupta2021calibration}, we divide the validation set of ImageNet into two halves: one for the in-distribution test (namely ImageNet-Val), the other for learning calibration methods (namely calibration set $\mathcal{D}_o$).

\textbf{Out-of-distribution test sets.} 
In the experiment, we use the following \emph{six real-world} out-of-distribution benchmarks. (i) \underline{ImageNet-V2} \cite{recht2019imagenet} is a new version of ImageNet test set. It contains three different sets resulting from different sampling strategies: Matched-Frequency (A), Threshold-$0.7$ (B), and Top-Images (C). Each version has $10,000$ images from $1000$ classes;
(ii) \underline{ImageNet-S(ketch)} \cite{wang2019learning} shares the same $1000$ classes as ImageNet but all the images are black and white sketches. It contains $50,000$ images;
(iii) \underline{ImageNet-R(endition)} \cite{hendrycks2021many} contains artificial renditions of ImageNet classes. It has $30,000$ images of $200$ classes. Following \cite{hendrycks2021many}, we sub-select the model logits for the $200$ classes before computing calibration metrics.
(iv) \underline{ImageNet-Adv(ersarial)} \cite{hendrycks2021nae} is adversarially selected to be hard for ResNet-$50$ trained on ImageNet. It has $7,500$ samples of $200$ classes. As for ImageNet-R, we sub-select the logits for the $200$ classes before computing the calibration metric. 
Moreover, we test on synthetic \underline{CIFAR-$10$-C(orruptions)} and \underline{ImageNet-C(orruptions)} \cite{hendrycks2019robustness}.
Both these two datasets are modified with synthetic perturbations such as blur, pixelation, and compression artifacts at a range of severities. We use $80$ different distortions ($16$ different types with $5$ levels of intensity each) which are the same as those in \cite{ovadia2019can}.

\textbf{Post-hoc calibration methods.} In the experiment, we validate the effectiveness of ACE by integrating it with the existing calibration methods through which we obtain calibrated logits $\mathbf{z}$ (Section \ref{method:ensemble}). Specifically, we use vector scaling \cite{guo2017calibration}, temperature Scaling \cite{guo2017calibration}, and Spline \cite{gupta2021calibration} as baseline calibrators, and compare with a recent method Perturbation \cite{tomani2021post} which is specifically designed for OOD calibration.
In addition, we also compare with more existing methods, \emph{i.e.,} Ensemble \cite{lakshminarayanan2017simple,ovadia2019can}, SVI \cite{wen2018flipout}, SVI-AvUC and SVI-AvUTS \cite{krishnan2020improving}, to show our method competitive. 

\setlength{\tabcolsep}{8pt}
\begin{table*}[t]
\small
\caption{OOD calibration performance (ECE, \%) of our method (ACE) and Perturbation \cite{tomani2021post} applied on Spline \cite{gupta2021calibration}. We report results using three neural networks: ResNet-$152$ \cite{he2016deep} (ResNet), ViT-Small-Patch$32$-$224$ \cite{dosovitskiy2020image} (ViT), and Deit-Small-Patch$16$-$224$ \cite{deit} (Deit).
All the other notations and settings are the same with Table \ref{resnet}. Our method improves the calibrator baselines in $16$ out of $18$ scenarios, while Perturbation has mixed performance.
}
\begin{center}
\begin{tabular}{llccccccc}
\toprule
 Models & Methods &  ImgNet-V2-A &  ImgNet-V2-B &  ImgNet-V2-C &  ImgNet-S &  ImgNet-R &  ImgNet-Adv\\
\midrule
\multirow{3}{*}{ResNet}
& Spline& $\underline{4.5321}$ & $\textbf{1.8034}$ & $\underline{1.3357}$ & $19.6392$  & $13.1116$ & $45.3623$\\
& + ACE & $\textbf{2.8201}$ \textcolor{green}{$\blacktriangle$}& $\underline{2.0235}$ \textcolor{red}{$\blacktriangledown$}& ${\textbf{1.0550}}$ \textcolor{green}{$\blacktriangle$}& $\textbf{6.9264}$ \textcolor{green}{$\blacktriangle$}& $\underline{6.8533}$ \textcolor{green}{$\blacktriangle$}& $\textbf{31.0926}$ \textcolor{green}{$\blacktriangle$}\\
& + Perturbation & $5.4175$ \textcolor{red}{$\blacktriangledown$}& $8.2109$ \textcolor{red}{$\blacktriangledown$}& $9.3326$ \textcolor{red}{$\blacktriangledown$}& $\underline{7.9805}$ \textcolor{green}{$\blacktriangle$} & $\textbf{2.9171}$ \textcolor{green}{$\blacktriangle$} & $\underline{32.3677}$ \textcolor{green}{$\blacktriangle$} \\
\midrule

\multirow{3}{*}{ViT}
& Spline & $\underline{4.7572}$ & $\textbf{1.6859}$ & $\underline{1.4683}$ & $15.9864$ & $12.5494$ & $38.0404$ \\
& + ACE & $\textbf{2.9329}$ \textcolor{green}{$\blacktriangle$}& $\underline{2.0832}$ \textcolor{red}{$\blacktriangledown$}& $\textbf{1.1831}$ \textcolor{green}{$\blacktriangle$}&$\textbf{4.8514}$ \textcolor{green}{$\blacktriangle$}& $\underline{6.3699}$ \textcolor{green}{$\blacktriangle$}& $\underline{23.5147}$ \textcolor{green}{$\blacktriangle$}\\
&  + Perturbation & $5.0302$ \textcolor{red}{$\blacktriangledown$}& $6.3854$ \textcolor{red}{$\blacktriangledown$}& $7.8929$ \textcolor{red}{$\blacktriangledown$}& $\underline{5.9254}$ \textcolor{green}{$\blacktriangle$} &$\textbf{3.7302}$ \textcolor{green}{$\blacktriangle$} & $\textbf{22.5118}$ \textcolor{green}{$\blacktriangle$}\\
\midrule

\multirow{3}{*}{Deit}
& Spline & $5.0289$ & $\underline{2.1261}$ & $\underline{1.3923}$ & $20.7714$ & $9.6996$ & $31.3674$\\
& + ACE & $\textbf{2.4576}$ \textcolor{green}{$\blacktriangle$}& $\textbf{1.6475}$ \textcolor{green}{$\blacktriangle$}& $\textbf{1.3544}$ \textcolor{green}{$\blacktriangle$}& $\textbf{5.6622}$ \textcolor{green}{$\blacktriangle$}& $\textbf{3.6721}$ \textcolor{green}{$\blacktriangle$}& $\textbf{15.7885}$ \textcolor{green}{$\blacktriangle$}\\
& + Perturbation & $\underline{3.3520}$ \textcolor{green}{$\blacktriangle$}& $2.4547$ \textcolor{red}{$\blacktriangledown$}& $2.9461$ \textcolor{red}{$\blacktriangledown$}& $\underline{15.9003}$ \textcolor{green}{$\blacktriangle$}& $\underline{8.1481}$ \textcolor{green}{$\blacktriangle$}& $\underline{27.9474}$ \textcolor{green}{$\blacktriangle$}\\
\bottomrule
\end{tabular}\label{vit}
\end{center}
\vskip -0.1in
\end{table*}

\setlength{\tabcolsep}{8pt}
\begin{table*}[ht]
\begin{center}
\begin{small}
\caption{Method comparison on CIFAR-$10$-C and ImageNet-C with ResNet-$20$ and ResNet-$50$, respectively. Following the protocol in~\cite{ovadia2019can}, we report mean ECE ($10$ bins for CIFAR-$10$-C and $25$ bins for ImageNet-C, $\%$) across 16 different types of data shift at intensity $5$ with lowest numbers in \textbf{bold} and the second lowest \underline{underlined}.
For each row, we compare ACE with the best of the competing ones (\emph{i.e.,} SVI-AvUC) using the two-sample t-test.
}
\begin{tabular}{lccccccc}
\toprule
Dataset & Uncalibrated & Ensemble \cite{lakshminarayanan2017simple} & SVI \cite{wen2018flipout} & SVI-AvUTS \cite{krishnan2020improving} & SVI-AvUC & Spline \cite{gupta2021calibration} & Spline+ACE \\
\midrule
CIFAR-10-C & $0.1942$ & $0.1611$ & $0.2389$ & $0.1585$ & $\underline{0.1374}$ & $0.3382$ & $\textbf{0.1272}$ \textcolor{green}{$\blacktriangle$}\\
\midrule
ImageNet-C & $0.3151$ & $0.0880$ & $0.1188$ & $0.0800$ & $\underline{0.0542}$ & $0.1147$ & $\textbf{0.0477}$ \textcolor{green}{$\blacktriangle$} \\
\bottomrule
\end{tabular}
\label{sota2}
\end{small}
\end{center}
\vskip -0.2in
\end{table*}

\subsection{Calibration on Out-of-distribution Datasets}
\label{exp:ood}
\textbf{ACE improves calibration methods on OOD datasets}. 
We evaluate our method combined with three post-hoc calibrators on six out-of-distribution test sets and compare it with those calibrators used alone.
Table \ref{resnet} shows ECE (using $25$ bins) results of ResNet-$152$. 
Our ACE is shown to consistently improve the OOD calibration results of the three baseline calibrators in most of the test cases. For example, when calibrating ResNet-152, our method improves temperature scaling by $0.88\%$, $0.17\%$, $5.40\%$, $3.73\%$ and $4.57\%$ decrease in ECE, on ImageNet-V2-A/B, ImageNet-S/R/Adv, respectively. Under the same settings, the ECE of our method is slightly higher ($0.019\%$) than the baseline on the ImageNet-V2-C dataset. 
We also report other metrics (\emph{e.g.,} Brier Score, KS-Error) in the supplemental material.

\textbf{ACE works effectively under two other neural networks}. To show the effectiveness of our method for different backbones, we adopt two transformer models (ViT-Small-Patch$32$-$224$ and Deit-Small-Patch$16$-$224$) as backbones, and experimental settings are the same as those in Table \ref{resnet}. Table \ref{vit} indicates that for backbone ViT-Small-Patch$32$-$224$ our method reduces ECE of the three baselines on five out of the six OOD test sets. %
For example, compared with Spline, ECE of our method is $1.82\%$, $0.28\%$, $11.13\%$, $6.16\%$ and $14.53\%$ lower on ImageNet-V2-A/C, ImageNet-S/R/Adv, respectively.
On the other hand, Table \ref{vit} demonstrates that for the Deit-Small-Patch$16$-$224$ backbone, our method is beneficial on all the six OOD test sets.
In addition, comparing the \emph{uncalibrated} results of the three backbones, transformer models generally have a lower ECE under OOD test sets. Specifically, \emph{ViT-Small-Patch$32$-$224$} is shown to be superior to Deit-Small-Patch$16$-$224$ on four out of six test sets. 

\textbf{Comparison with the existing calibration methods.}
In Table \ref{sota2}, we compare our method with the state-of-the-art methods, \emph{i.e.}, various variants of AvUC \cite{krishnan2020improving} and Ensemble \cite{lakshminarayanan2017simple}, on CIFAR-$10$-C and ImageNet-C. Following the protocol in  \cite{ovadia2019can, krishnan2020improving}, we report the results at intensity $5$.
Our method improves Spline by reducing ECE by $11.18\%$ and $6.70\%$  on CIFAR-$10$-C and ImageNet-C, respectively. Compared with these methods, our method is competitive on both ImageNet-C and CIFAR-$10$-C. For example, for CIFAR-$10$-C, our method achieves $3.21\%$ and $1.10\%$ lower calibration error than SVI-AvUTS and SVI-AvUC, respectively. 

\begin{figure}[t]
    \centering
    \centerline{\includegraphics[width=1.0\columnwidth]{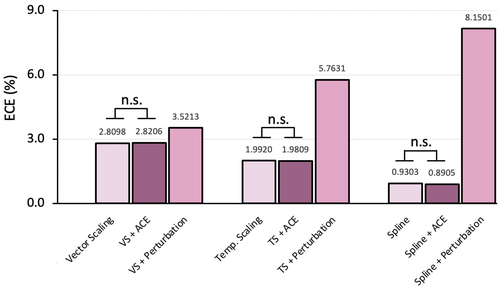}}
    \caption{\textbf{Evaluation of ACE on the ID test set ImageNet-Val.} We calibrate the ResNet-$152$ classifier and use ECE (\%) for top-$1$ predictions as evaluation metric. ``n.s." means  the difference between results is not statistically significant ($p$-value $\textgreater 0.05$).
    }\label{ID}
    \vskip -0.25in
\end{figure}

\subsection{ACE Does Not Compromise ID Calibration}\label{sec:in-distribution}
We show ECE results on in-distribution test set (ImageNet-Val) using ResNet-$152$. We adopt the same three post-hoc calibration baselines and Perturbation \cite{tomani2021post} for comparison.
As shown in Fig. \ref{ID}, we observe that the post-hoc calibration baselines themselves effectively reduce the ECE score compared with the uncalibrated system and that Spline generally performs the best.
Perturbation is shown to deteriorate the calibration performance for all three baselines. 
Because Perturbation is not adaptive to different test sets, its effectiveness is not guaranteed when a test set is out of its optimal domain confined by the generated diverse set.
In comparison, when our method is integrated with the baselines, the resulting calibration performance is very close to the baselines when being used alone. This is mainly because of the adaptive weighting scheme (see Section \ref{method:ensemble} for more explanations). Thus our method is \emph{not} compromised on the in-distribution test set.

\begin{figure}[t]
    \centering
    \centerline{\includegraphics[width=0.9\columnwidth]{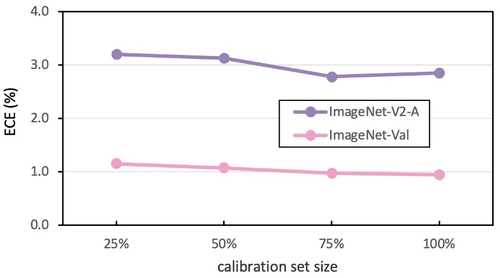}}
    \caption{
    \textbf{Effect of the size of the two extreme calibration sets.} Starting from original size ($25,000$ and $5,885$ images respectively for $\mathcal{D}_o$ and $\mathcal{D}_h$), we randomly select a certain percentage of calibration sets. We report ECE of ResNet-$152$ with Spline on ImageNet-Val and ImageNet-V2-A.
    }\label{size}
    \vskip -0.25in
\end{figure}

\subsection{Component Analysis of ACE}\label{sec:component}

\textbf{Impact of the size of the two extreme calibration sets.} %
ACE uses an ``easy'' calibration set $\mathcal{D}_o$ (the original calibration set) and a ``hard'' calibration set $\mathcal{D}_h$. The original $\mathcal{D}_o$ and $\mathcal{D}_h$ have $25,000$ and $5,885$ images, respectively. Here, we simultaneously reduce the size of $\mathcal{D}_o$ and $\mathcal{D}_h$ by a certain percentage and report calibration performance (ECE) in Fig. \ref{size}. From the results on the in-distribution dataset ImageNet-Val and out-of-distribution dataset ImageNet-V2-A, we observe that our method is relatively stable on both test sets when the size changes. Yet for best results, we recommend using possibly large calibration sets.

\begin{figure}[t]
\begin{center}
\centerline{\includegraphics[width=0.85\columnwidth]{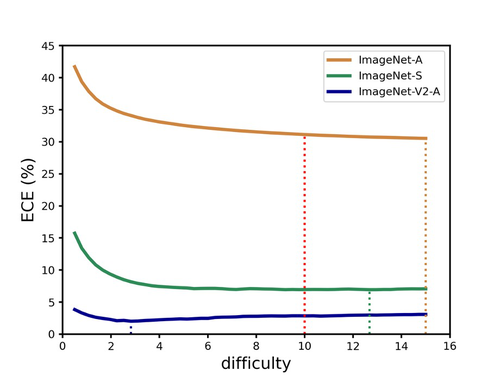}}
\caption{
\textbf{Impact of hyperparameter $d$ on OOD calibration.} We densely sample values of $d\in(0.5,15)$ and report ECE (\%) of ResNet-$152$ with Spline on ImageNet-V2-A, ImageNet-S and ImageNet-Adv.
We also mark the results using our empirically selected value ($d=10$) and the optimal values shown by the dotted vertical line with the same color.
}
\vskip -0.45in
\label{d_h}
\end{center}
\end{figure}

\begin{figure}[t]
    \centering
    \includegraphics[width=0.85\columnwidth]{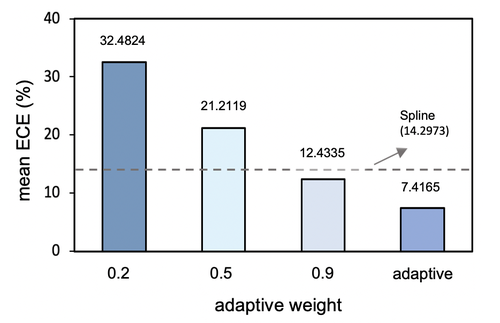}
    \caption{
    \textbf{Comparison of different weighting schemes for ACE.}
    We report the mean ECE (\%) on six OOD datasets (ImageNet-V2-A/B/C, ImageNet-S, ImageNet-Adv, ImageNet-R) and one ID test set (ImageNet-Val). Spline and ResNet-$152$ is used.
    }
    \vskip -0.25in
    \label{fig:ad-hoc}
\end{figure}

\textbf{Impact of the difficulty of $\mathcal{D}_h$.} To analyze the impact of hyperparameter $d$ (Section \ref{method:ensemble}), we create multiple $\mathcal{D}_h$ with various values of $d$. Results are shown in Fig. \ref{d_h}. 
We observe that calibration performance is slightly higher on ImageNet-Adv when the hard calibration set is more difficult, while the performance on the other two datasets drops at the same time. Moreover, we find the optimal difficulty is different for various test sets. That said, by setting $d=10$, we generally have good performance, and it is important to note that this difficulty level is considerably high (equivalent to 9.09\% classification accuracy) and thus covers most test scenarios.

\textbf{Comparing fixed weighting schemes with the adaptive weight}
We compare the adaptive weight ($\alpha=\frac{\text{avgConf}(\mathcal{D}_{test})}{ \text{avgConf}(\mathcal{D}_{o})}$) with setting $\alpha$ to a few fixed values $0.2$, $0.5$, and $0.9$. 
The difficulty level for the out-of-distribution situation is $10$. We evaluate the three calibration baselines on the six out-of-distribution test sets and one in-distribution test set using ResNet-$152$ as backbone and use the mean ECE ($\%$) value over all the seven test sets (six OOD datasets and one ID test set) as evaluation metric.

As shown in Fig. \ref{fig:ad-hoc}, when applying our ACE on Spline, using $\alpha=0.2$ and $\alpha=0.5$ deteriorate calibration performance, while $\alpha=0.9$ improves the baseline. 
However, without test labels, it is infeasible to set an appropriate $\alpha$ for each test set. Moreover, the test sets are changed, setting fixed values of $\alpha$ might be effective in some cases and be less useful in others. 
In contrast, our designed test-set-adaptive $\alpha$ ( Eq. 
\ref{eq:weight}) is shown to improve the baselines on various OOD test sets.
Also, we report the value of $\alpha$ used for each test set in Table \ref{resnet} and Table \ref{vit} in the supplemental material.

\textbf{Optimal adaptive $\alpha$ vs. our computed $\alpha$ (Eq. \ref{eq:weighted_sum}).} We compare both values in Fig. \ref{alpha}.
First, for datasets with normal difficulty (\emph{e.g.,} ImageNet-V2-A), the value computed by our scheme is quite close to the optimal value.
Second, for extremely difficult datasets such as ImageNet-S and ImageNet-A, $\alpha$ computed by our proposed method is less optimal. That said, we emphasize that in practice it is infeasible to do a greedy search because the images of test set are unlabeled, where our ACE method is generally useful. 

\begin{figure}[t]
\begin{center}
\centerline{\includegraphics[width=0.85\columnwidth]{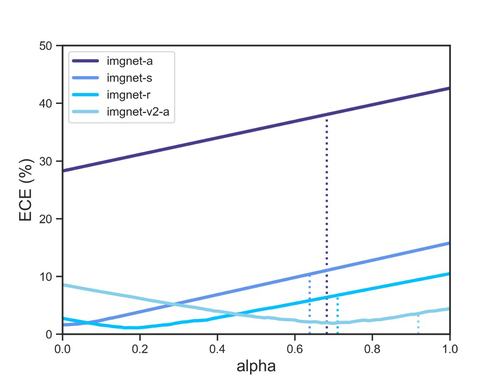}}
\caption{
\textbf{Densely sampled values of $\alpha$ (0 to 1) \emph{vs.} our computed $\alpha$ (Eq. \ref{eq:weight})} Comparing with the densely sampled values of $\alpha$, computed $\alpha$ (shown by the dotted vertical line) is close to the optimal value with reasonable difficulty for each test set.
}
\vskip -0.5in
\label{alpha}
\end{center}
\end{figure}

\section{Conclusion}\label{sec:conclusion}

This paper studies how to calibrate a model on OOD datasets. 
Our important contribution is diagnosing why existing post-hoc algorithms fail on OOD test sets. Specifically, we report the difficulty of the calibration set influences the calibration function learning, and in other words, an OOD test set would witness poor calibration performance if the calibration set does not have an appropriate difficulty level. 
Realizing the importance of calibration set difficulty, we design a simple and effective method named adaptive calibrator ensemble (ACE) which combines the outputs of two calibrators trained on datasets with extreme difficulties.
We also demonstrate how the ensemble scheme works for temperature scaling.
We show that ACE improves three commonly used calibration methods on various OOD calibration benchmarks (\emph{e.g.}, ImageNet-C and CIFAR-10-C) without degrading ID calibration performance. In future work, we would like to further study how the domain gap and calibration set difficulty interact with each other and thereby improve OOD calibration.

\clearpage

{\small
\bibliographystyle{ieee_fullname}
\bibliography{egbib}
}
\clearpage
\appendix
\setcounter{table}{0}
\renewcommand{\thetable}{A\arabic{table}}
\setcounter{table}{0}
\renewcommand{\thetable}{A\arabic{table}}

We first introduce the experimental setup including training details, dataset split, and computation resources. We also report more metrics (\emph{i.e.,} KSE \cite{gupta2021calibration} and BS \cite{brier1950verification}) in Table \ref{table:metric} and detailed statistical test results of Table 1 in main paper.
Then, we provide more comparative results with Perturbation \cite{tomani2021post} in Table \ref{table:pert}, and we report full results on CIFAR-$10$-C and ImageNet-C in Table \ref{CIFAR10C} and Table \ref{ImageNetC}, respectively. 
Lastly, we give more component analysis of the proposed ACE method in Section \ref{sec:component}.

\section{Experimental Setup}

\subsection{CIFAR-10 Setup}
Following the protocol in \cite{guo2017calibration,krishnan2020improving}, we use $5,000$ images from the training set of CIFAR-$10$ as the calibration set. We use ResNet-$20$ designed for CIFAR-$10$ and train it using publicly available codes in \cite{krishnan2020improving}.

\subsection{ImageNet Setup} 
Following the protocol in \cite{guo2017calibration}, we divide the validation set of ImageNet into two halves: one for in-distribution test; the other for learning calibration methods. We use ResNet-$50$, ResNet-$152$, Vit-Small-Patch$32$-$224$ and Deit-Small-Patch$16$-$224$. Their weights are publicly provided by PyTorch Image Models (timm-$0.5.4$) \cite{rw2019timm}. 

\subsection{Baseline Methods}
Our proposed ACE method is used for improving post-hoc methods (\emph{i.e.}, Vector Scaling, Temperature Scaling, and Spline) on OOD test sets. 
For each baseline, we use the publicly available codes to train the calibration model. We follow the code and use the same training settings (such as regularization, training scheduler, and training hyper-parameters). The codes we used are:
\\
\textbf{Vector Scaling}: \\
\textcolor{blue}{https://github.com/saurabhgarg1996/calibration}
\\
\textbf{Temperature Scaling}: \\
\textcolor{blue}{https://github.com/gpleiss/temperature\_scaling}
\\
\textbf{Spline}: \\
\textcolor{blue}{https://github.com/kartikgupta-at-anu/spline-calibration}

\subsection{More Metrics for Table 1}

We report the ECE ($\%$) result in Table 1. To better prove the effectiveness of our method, we report another two classic metrics: KSE ($\%$) \cite{gupta2021calibration}
and Brier Score ($\%$) in Table \ref{table:metric}. 
The results in table \ref{table:metric} shows that our method is also effective with these metrics. 

\subsection{The Statistical Significance Test in Table 1}

We adopt the two-sample t-test, which tells whether the performance of the baseline and baseline + ACE has a significant difference.  All methods are run for $5$ times based on $5$ random seeds ($1,2,3,4,5$).

Given a random seed, we use it to randomly downsample the hard calibration set from the original validation set. 
For all random seeds, the samples for the baseline are indeed the same. However, when training a calibrator, every mini-batch is randomly sampled and shuffled, thus resulting in randomness.
As reported in Table \ref{sst1} of the main paper (mean and standard deviation of ECE), the impact of different random seeds is slight.
We also adopt the Welch’s t-test in Table \ref{sst2} to validate this.

\subsection{Computation Resource} 
We use the Pytorch-$1.9.1$ framework and run all the experiment on one GPU (GeForce RTX $2080$ Ti). The CPU is $24$  Intel(R) Core(TM) i$9$-$10920$X CPU @ $3.50G$Hz.

\subsection{Datasets} 
We list the links of the used datasets and check carefully their licenses for our usage.
\\
\textbf{ImageNet-Validation} \cite{deng2009imagenet} (\textcolor{blue}{https://www.image-net.org}); \\
\textbf{ImageNet-V2-A/B/C} \cite{recht2019imagenet} \\
(\textcolor{blue}{https://github.com/modestyachts/ImageNetV2}); \\
\textbf{ImageNet-Corruption} \cite{hendrycks2019robustness} \\
(\textcolor{blue}{https://github.com/hendrycks/robustness});\\
\textbf{ImageNet-Sketch} \cite{wang2019learning} \\
(\textcolor{blue}{https://github.com/HaohanWang/ImageNet-Sketch});\\
\textbf{ImageNet-Adversarial} \cite{hendrycks2021nae} \\
(\textcolor{blue}{https://github.com/hendrycks/natural-adv-examples});\\
\textbf{ImageNet-Rendition} \cite{hendrycks2021many} \\
(\textcolor{blue}{https://github.com/hendrycks/imagenet-r});\\
\textbf{CIFAR-$10$} \cite{krizhevsky2009learning}(\textcolor{blue}{https://www.cs.toronto.edu/~kriz/cifar.html});\\
\textbf{CIFAR-$10$-C} \cite{hendrycks2019robustness}(\textcolor{blue}{https://github.com/hendrycks/robustness}); \\

\section{More Comparison}
\subsection{Comparison with Perturbation}
In Table \ref{sota}, we compare our method with a recent OOD calibration method Perturbation \cite{tomani2021post}. In Table \ref{sota}, we observe that Perturbation improves the baselines on Level $5$ of ImageNet- C. In fact, these test sets contain data that are seriously out of distribution. However, for datasets that lean towards being in-distribution, \emph{e.g.}, Level $1$ in ImageNet-C, Perturbation worsens the baselines. A probable reason is that the diverse calibration set where Perturbation is trained is closer to heavily OOD data (Level-$5$). In comparison, our method adapts to various test sets through the weighting scheme and yields improvement with statistical significance in most test cases.

\clearpage

\setlength\tabcolsep{2.5pt}
\begin{table*}[t]
\begin{center}
\caption{We used two other metrics, Brier Score ($\%$), KS-Error ($\%$) \cite{gupta2021calibration}.
We evaluate two calibrators (Temperature Scaling and Spline). All other settings remain the same with Table 1
}
\label{table:metric}
\vskip 0.05in
\begin{small}
\begin{tabular}{lccccccc}
\toprule
 Metric & Methods & ImgNet-V2-A &  ImgNet-V2-B &  ImgNet-V2-C &  ImgNet-S &  ImgNet-R &  ImgNet-Adv\\
\midrule
\multirow{5}*{KSE} & UnCal & $5.2260$ & $9.5910$ & $4.0399$ & $24.6331$ & $17.8626$ & $50.8544$\\
& Temp.Scaling & $4.0937$ & $1.1129$ & $0.8773$ & $15.7880$ & $10.4752$ & $42.6302$\\
& +ACE & \underline{$3.0661$} & 
\underline{$0.7809$} & \underline{$0.8406$} & $\textbf{1.0386}$ & \underline{$6.7335$} & \underline{$38.0691$}\\
& Spline & $4.4217$ & $1.0765$ & $0.8813$ & $19.6394$ & $13.0808$ & $45.3623$\\
& +ACE & $\textbf{1.2029}$ & $\textbf{0.7239}$ & $\textbf{0.3483}$ & \underline{$5.8538$} & $\textbf{3.5370}$ & $\textbf{31.1308}$\\
\midrule
\multirow{5}*{BS} & UnCal & $15.7902$ & $13.0527$ & $11.1197$ & $21.6672$ & $18.0285$ & $39.1104$\\
& Temp.Scaling & $14.8083$ & $12.6830$ & $10.9561$ & $17.2627$ & $15.2080$ & $30.3974$\\
& +ACE & $\underline{14.7192}$ & $12.6815$ & $10.9532$ & $\underline{15.3793}$ & $\textbf{14.3487}$ & $\underline{26.2166}$\\
& Spline & $14.8779$ & $\textbf{12.5798}$ & $\underline{10.8702}$ & $18.9953$ & $16.1986$ & $32.0494$\\
& +ACE & $\textbf{14.7086}$ & $\underline{12.5804}$ & $\textbf{10.8640}$ & $\textbf{14.9486}$ & $\underline{14.6938}$ & $\textbf{18.8537}$\\

\bottomrule
\end{tabular}\label{sst1}
\end{small}
\end{center}
\vskip -0.2in
\end{table*}

\setlength\tabcolsep{2.5pt}
\begin{table*}[!ht]
\begin{center}
\caption{The \emph{t-statistic} and \emph{$p$ values} of the two-sample t-test method in Table 1 of main paper. We report the resulting statistics and $p$ values here, which are one-on-one corresponded to the numbers in Table 1. We regard $p<0.05$ as statistically significant.
}
\label{table:pert}
\vskip 0.05in
\begin{small}
\begin{tabular}{lccccccc}
\toprule
 Methods & & ImgNet-V2-A &  ImgNet-V2-B &  ImgNet-V2-C &  ImgNet-S &  ImgNet-R &  ImgNet-Adv\\
\midrule
\multirow{2}*{Vector Scaling} & t-statistic &$59.25$ &$37.39$&$-25.14$&$355.60$&$170.03$&$217.22$\\
& $p$ & $7.31e^{-12}$&$2.87e^{-10}$&$6.70e^{-9}$&$4.37e^{-18}$&$1.60e^{-15}$&$2.25e^{-16}$
\\

\midrule
\multirow{2}*{Temp. Scaling}& t-statistic& $615.89$&$249.42$&$-195.10$&$1164.86$&$800.82$&$898.46$ \\
&$p$& $5.40e^{-20}$&$7.47e^{-17}$&$5.33e^{-16}$&$3.30e^{-22}$&$6.62e^{-21}$&$2.63e^{-21}$\\
\midrule
\multirow{2}*{Spline}& t-statistic&$120.74$&$-28.46$&$60.99$&$294.01$&$675.16$&$109.61$
\\
 &$p$&$2.47e^{-14}$&$2.50e^{-9}$&$5.80e^{-12}$&$2.00e^{-17}$&$2.59e^{-20}$&$5.36e^{-14}$
 \\

\bottomrule
\end{tabular}\label{sst1}
\end{small}
\end{center}
\vskip -0.2in

\end{table*}

\setlength\tabcolsep{2.5pt}
\begin{table*}[!ht]
\begin{center}
\caption{The \emph{t-statistic} and \emph{$p$ values} of the Welch's t-test in Table 1 of main paper. We report the resulting statistics and $p$ values here, which are one-on-one corresponded to the numbers in Table 1. We regard $p<0.05$ as statistically significant.
}
\label{table:pert}
\vskip 0.05in
\begin{small}
\begin{tabular}{lccccccc}
\toprule
 Methods & & ImgNet-V2-A &  ImgNet-V2-B &  ImgNet-V2-C &  ImgNet-S &  ImgNet-R &  ImgNet-Adv\\
\midrule
\multirow{2}*{Vector Scaling} & t-statistic &$59.25$ &$37.39$&$-25.14$&$355.60$&$170.03$&$217.22$\\
& $p$ & $4.85e^{-7}$&$3.05e^{-6}$&$1.48e^{-5}$&$3.75e^{-10}$&$7.17e^{-9}$&$2.68e^{-9}$
\\

\midrule
\multirow{2}*{Temp. Scaling}& t-statistic& $615.89$&$249.42$&$-195.10$&$1164.86$&$800.82$&$898.46$ \\
&$p$& $4.16e^{-11}$&$1.55e^{-9}$&$4.14e^{-9}$&$3.25e^{-12}$&$1.45e^{-11}$&$9.20e^{-12}$\\
\midrule
\multirow{2}*{Spline}& t-statistic&$120.74$&$-28.46$&$60.99$&$294.01$&$675.16$&$109.61$
\\
 &$p$&$2.82e^{-8}$&$9.06e^{-6}$&$4.32e^{-7}$&$8.02e^{-10}$&$2.88e^{-11}$&$4.15e^{-8}$
 \\

\bottomrule
\end{tabular}\label{sst2}
\end{small}
\end{center}
\vskip -0.2in

\end{table*}

\setlength\tabcolsep{4pt}
\begin{table*}[t]
\begin{center}
\caption{Method comparison on \textbf{ImageNet-C} datasets \cite{hendrycks2019robustness}. We report ECE (\%) for top-$1$ predictions (in \%) of the ResNet-$152$ model.
For each level of corruption (column), we report the average ECE using $25$ bins with lowest numbers in \textbf{bold} and second lowest \underline{underlined}.
ACE improves calibration performance of two post-hoc calibration methods on all datasets.
}
\label{table:pert}
\begin{tabular}{lccccc}
\toprule
 & \multicolumn{5}{c}{Corruption Intensity}\\
\cmidrule(r){2-6}
 Method & Level $1$ & Level $2$ & Level $3$ & Level $4$ & Level $5$\\
\midrule
Uncalibrated & $6.0684$ & $7.8617$ &$9.7938$&$12.3911$&$15.5049$\\
\midrule
Temperature Scaling (TS) &$\underline{2.4880}$ & $\textbf{2.7976}$& $\underline{3.7996}$ &$5.1836$ &$7.7213$\\
Temperature + Perturbation &$9.3084$ &$8.6574$ &$7.6707$ &$5.7594$ & ${4.3672}$\\
Temperature + ACE &$2.9733$& $\underline{3.1130}$& $\textbf{3.1306}$&$\textbf{3.1494}$ &$\underline{4.3034}$\\
\midrule
Spline &$\textbf{1.8049}$ & $3.1690$& $5.2388$& $7.8672$&$11.0547$\\
Spline + Perturbation & $9.6207$&$8.1570$ &$6.7643$ &$5.1064$ & $5.2777$\\
Spline + ACE &$3.6982$& $4.2046$&$4.2944$ &$\underline{3.7231}$ &$\textbf{3.9707}$\\
\bottomrule
\end{tabular}\label{sota}
\end{center}
\end{table*}

\clearpage

\begin{figure*}[t]
    \begin{minipage}{0.5\linewidth}
    \centering
    {\includegraphics[width=0.9\columnwidth]{Figs/ad-hoc-alpha.png}}
    \end{minipage}
    \begin{minipage}{0.45\linewidth}
    \centering
    {\includegraphics[width=0.9\columnwidth]{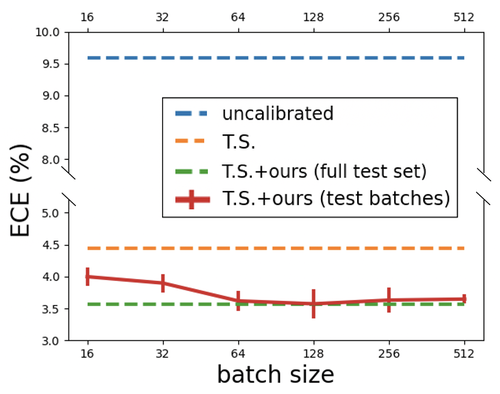}}
    \end{minipage}
    \caption{
    \textbf{Left}: 
    Comparison of different weighting schemes for ACE. We report the mean ECE (\%) on six OOD datasets (ImageNet-V2-A/B/C, ImageNet-S, ImageNet-Adv, ImageNet-R) and one ID test set (ImageNet-Val). Spline is used as the calibration baseline. The ResNet-$152$ model is used.
    \textbf{Right}: 
    Effectiveness of $\alpha$ computed in test batches of different sizes ($16$, $64$, $128$, $256$ and $512$). Comparing with calculating $\alpha$ on the full set, using test batches yields similar ECE (\%) especially when the batch size is at least $64$. Temperature scaling (T.S.) is used as the baseline calibrator for our ACE. We also include the original baseline results in the figure.
    }
    \label{fig1}
\end{figure*}

\begin{table*}[!ht]
\begin{center}
\caption{Method comparison on \textbf{ImageNet-V2-A, ImageNet-V2-B, ImageNet-V2-C, and ImageNet-S} datasets. Following the protocol in \cite{wang2020transferable}, we report ECE (\%) for top-$1$ predictions (in \%) of the ResNet-$50$ model.
}
\label{table:pert}
\begin{tabular}{lcccc}
\toprule
Method & ImageNet-V2-A & ImageNet-V2-B & ImageNet-V2-C & ImageNet-S\\
\midrule
Uncalibrated & $9.50$ & $6.23$ & $4.31$ &$22.32$ \\
Temperature Scaling & $4.44$ & $2.73$ & $1.68$ & $16.27$ \\ 
TransCal &$12.26$ &$4.43$ & $1.86$ & $8.10$ \\ 
Ours &$3.56$ &$2.56$ &$1.70$ & $7.53$\\
\bottomrule
\end{tabular}\label{transcal}
\end{center}
\end{table*}

\setlength\tabcolsep{3.2pt}
\begin{table*}[h]

    \centering
    \caption{Full results on CIFAR-$10$-C datasets  \cite{hendrycks2019robustness}. We report the lower quartile ($25$-th percentile), median ($50$-th percentile), mean and upper quartile ($75$-th percentile) of ECE computed across $16$ different types of data shift at intensity $5$ with lowest numbers in \textbf{bold} and second lowest \underline{underlined}.
    }
    \label{CIFAR10C}
    \begin{small}

    \begin{tabular}{ccccccccccc}
    \toprule
    \multirow{2}*{Metric} & & \multicolumn{9}{c}{Method} \\
    \cmidrule{3-11}
    & & Vanilla & \makecell[c]{Temp\\Scaling} & Ensemble & SVI & LL SVI & \makecell[c]{SVI \\ -AvUTS} & \makecell[c]{SVI \\ -AvUC} & Spline & \makecell[c]{Spline\\+Ours} \\
    \midrule
    \multirow{4}*{ECE} & lower quartile & {$0.2121$} & {$0.0997$} & {$0.0549$} & {$0.0925$} & {$0.2027$} & \underline{$0.0466$} & $\textbf{0.0398}$ & {$0.2045$} & {$0.0783$} \\
     & median quartile & $0.3022$ & $0.1834$ & $0.1054$ & $0.2146$& $0.3077$ & $0.1516$ & \underline{$0.1107$} & $0.3007$ & $\textbf{0.1071}$\\
     & mean &$0.3151$ & $0.1993$ & $0.1611$ & $0.2389$ & $0.3267$ & $0.1585$ & \underline{$0.1374$} & $0.3382$ & $\textbf{0.1272}$\\
     & upper quartile & $0.4148$ & $0.2915$ & $0.2551$ & $0.3636$ & $0.4246$ & $0.2345$ & \underline{$0.2303$} & $0.4376$ & $\textbf{0.1522}$\\
    \bottomrule
    \end{tabular}
    \end{small}
\end{table*}

\setlength\tabcolsep{3.2pt}
\begin{table*}[h]
    \centering
    \caption{Full results on ImageNet-C datasets  \cite{hendrycks2019robustness}. We report the lower quartile($25$-th percentile), median ($50$-th percentile), mean and upper quartile ($75$-th percentile) of ECE computed across $16$ different types of datashift at intensity $5$ with lowest numbers in \textbf{bold} and second lowest \underline{underlined}.
    }
    \label{ImageNetC}
    \begin{small}
    \begin{tabular}{ccccc ccccc c}
    \toprule
    \multirow{2}*{Metric} & & \multicolumn{9}{c}{Method} \\
    \cmidrule{3-11}
    & & Vanilla & \makecell[c]{Temp\\Scaling} & Ensemble & SVI & LL SVI & \makecell[c]{SVI\\-AvUTS} & \makecell[c]{SVI\\-AvUC} & Spline & \makecell[c]{Spline\\+Ours} \\
    \midrule
    \multirow{4}*{ECE} & lower quartile & $0.1244$ & $0.0959$ & $0.0503$ & $0.0722$ & $0.1212$ & $0.0420$ & \underline{$0.0319$} & $0.0575$ & $\textbf{0.0233}$\\
     & median quartile & $0.1737$ & $0.1392$ & $0.0900$ & $0.1144$& $0.1684$ & $0.0807$ & $\textbf{0.0447}$ & $0.1143$ & \underline{$0.0452$}\\
     & mean &$0.1942$ & $0.1600$ & $0.0880$ & $0.1188$ & $0.1868$ & $0.0800$ & \underline{$0.0542$} & $0.1147$ & $\textbf{0.0477}$\\
     & upper quartile & $0.2744$ & $0.2364$ & $0.1264$ & $0.1723$ & $0.2676$ & $0.1275$ & \underline{$0.0696$} & $0.1363$ & $\textbf{0.0606}$\\
    \bottomrule
    \end{tabular}
    \end{small}
\end{table*}

\clearpage

\setlength{\tabcolsep}{3pt}
\begin{table*}[t]
\begin{center}
\begin{small}
\caption{The adaptive $\alpha$ that we adopt in Table 1 and Table 2 of main paper.
}
\label{alpha}
\begin{tabular}{lccccccc}
\toprule
Model & ImgNet-Val & ImgNet-V2-A & ImgNet-V2-B & ImgNet-V2-C & ImgNet-S & ImgNet-R & ImgNet-Adv \\
\midrule
ResNet & $0.994080$ & $0.918328$ & $0.972311$ & $0.989697$ &$0.63765$&$0.709984$&$0.682187$ \\
Vit & $0.998655$& $0.896980$ & $0.969018$ & $0.98561$& $0.538366$ & $0.674307$ & $0.637850$ \\
Deit & $0.998741$ & $0.912270$ & $0.967555$ & $0.999048$ & $0.612748$ & $0.648445$ & $0.618136$ \\
\bottomrule
\end{tabular}
\label{est-c}
\end{small}
\end{center}
\end{table*}

\setlength{\tabcolsep}{3pt}
\begin{table*}[t]
\begin{center}
\begin{small}
\caption{Method comparison on CIFAR-$10$-C and ImageNet-C datasets with ResNet-$20$ and ResNet-$50$, respectively. Following the protocol in \cite{ovadia2019can}, we report mean ECE (\%) across 16 different types of data shift at intensity $5$ with lowest numbers in \textbf{bold} and second lowest \underline{underlined}. 
}
\label{estimate}
\begin{tabular}{lccccccc}
\toprule
Dataset & Vanilla & SVI & \makecell[c]{SVI\\-AvUC } & Spline & \makecell[c]{Spline\\+ACE} & \makecell[c]{Spline\\+Estimation} \\
\midrule
CIFAR-10-C & $0.1942$ & $0.2389$ & $0.1374$ & $0.3382$
& $\textbf{0.1264}$ & $\underline{0.1298}$\\
\midrule
ImageNet-C & $0.3151$ & $0.1188$ & $\underline{0.0542}$ & $0.1147$ & $\textbf{0.0477}$ & $0.0576$\\
\bottomrule
\end{tabular}
\label{est-c}
\end{small}
\end{center}
\end{table*}

\setlength{\tabcolsep}{3pt}
\begin{table*}[!ht]
\begin{center}
\begin{small}
\caption{Calibration performance of our method integrated with Temperature Scaling on one in-distribution test set and six OOD test sets. ECE (25bins, $\%$) for top-$1$ predictions. Here we $\mathcal{D}_o$  with the sample size of $\mathcal{D}_h$ ($5,884$).
}
\label{table:same_size}
\begin{tabular}{lccccccc}
\toprule
Method & ImgNet-Val & ImgNet-V2-A & ImgNet-V2-B & ImgNet-V2-C & ImgNet-S & ImgNet-R & ImgNet-Adv \\
\midrule
Temp.Scaling & $1.9670$ & $4.3571$ & $2.7234$ & $1.7880$ & $15.6735$ & $10.3832$ & $42.5225$\\
+ACE & $1.9623$ & $3.4842$ & $2.5458$ & $1.6764$ & $10.3131$ & $6.6726$ & $37.9957$\\
\bottomrule
\end{tabular}
\label{est-c}
\end{small}
\end{center}
\end{table*}

\setlength{\tabcolsep}{3pt}
\begin{table*}[t]
\begin{center}
\begin{small}
\caption{Calibration performance of our method integrated with Temperature Scaling on one in-distribution test set and six OOD test sets. ECE ($25$ bins, $\%$) for top-$1$ predictions. We use LCNet-$050$ and TinyNet-E, which have $60.094\%$ and $59.856\%$ top-$1$ accuracy, respectively on the validation set of ImageNet dataset. (Note IN is short for ImageNet)
}
\label{table:hard_do}
\begin{tabular}{lcccccccc}
\toprule
Model & Method & IN-Val & IN-V2-A & IN-V2-B & IN-V2-C & IN-S & IN-R & IN-Adv \\
\midrule
\multirow{2}*{LCENet-$050$} & Temp.Scaling & $1.8293$ & $6.6047$ & $2.9681$ & $1.6949$ & $20.3415$ & $18.9839$ & $43.1683$\\
& +ACE & $1.8238$ & $4.8591$ & $2.2639$ & $1.7516$ & $14.0055$ & $15.3397$ & $39.2584$\\
\midrule
\multirow{2}*{TinyNet-E} & Temp.Scaling & $1.3888$ & $6.8949$ & $2.7991$ & $1.7194$ & $22.4438$ & $20.7810$ & $41.3513$\\
& +ACE & $1.3857$ & $5.4262$ & $2.4606$ & $1.8311$ & $17.1741$ & $17.7259$ & $38.0800$\\
\bottomrule
\end{tabular}
\label{est-c}
\end{small}
\end{center}
\end{table*}

\setlength{\tabcolsep}{3pt}
\begin{table*}[t]
\begin{center}
\begin{small}
\caption{Calibration performance of our method integrated with Temperature Scaling and Spline on the in-distribution and OOD iWildCam-WILDS dataset. ECE (25bins, $\%$) for top-$1$ predictions and ResNet-$50$ classifier is used.
}
\label{table:wilds}
\begin{tabular}{lcccccc}
\toprule
Dataset & Uncal. & Temp.Scaling & Temp.Scaling+Ours & Spline & Spline+Ours \\
\midrule
iWildCam-WILDS-ID & $14.2701$ & $2.6786$ & $2.5833$ & $3.8142$ & $3.6965$\\
iWildCam-WILDS-OOD & $13.5552$ & $4.8231$ & $3.9738$ & $4.9902$ & $4.8425$\\
\bottomrule
\end{tabular}
\end{small}
\end{center}
\end{table*}

\setlength{\tabcolsep}{3pt}
\begin{table*}[t]
\begin{center}
\begin{small}
\caption{Calibration performance of different combination schemes. ECE (25bins, $\%$) for top-$1$ predictions is reported. Spline baseline and ResNet-$152$ classifier is used.
}
\label{table:comb}
\begin{tabular}{lcccccc}
\toprule
Method & ImageNet-V2-A & ImageNet-V2-B & ImageNet-V2-C & ImageNet-S & ImageNet-R & ImageNet-Adv\\
\midrule
Uncal. & 9.5016 & 6.2311 & 4.3117 & 24.6332 & 17.8621 & 50.8544\\
$\mathbf{z}_o^{\alpha} \otimes \mathbf{z}_h^{1-\alpha}$ & 5.0091 & 2.7478 &1.3357 & 6.4506 & 10.2066 & 28.4341 \\
$\alpha\cdot\mathbf{z}_o + (1-\alpha)\cdot\mathbf{z}_h$ & 2.8201 & 2.0235 &1.0550 & 6.9264 & 6.8533 & 31.0926 \\
\bottomrule
\end{tabular}
\label{table:comb}
\end{small}
\end{center}
\end{table*}

\clearpage

\subsection{Comparison with TransCal}
In Table \ref{transcal}, we compare our method with a recent OOD calibration method TranCal \cite{wang2020transferable}. In Table \ref{transcal}, we observe that TransCal is inferior to our method on the ImageNet-S dataset with ResNet-$50$. 

\section{Full Results on ImageNet-C and CIFAR-10-C}
In the Table 3 of the main paper, we report the mean ECE ($\%$) across $16$ different types of data shift at intensity $5$. In addition, we report the complete ECE results on CIFAR-$10$-C and ImageNet-C at intensity $5$ in Table \ref{CIFAR10C} and Table \ref{ImageNetC}. We observe that our method effectively improves the baselines (Spline) and gives state-of-the-art calibration accuracy under $2$ out of $3$ quartiles and mean value on both CIFAR-$10$-C and ImageNet-C.

\section{More Component Analysis}
\label{sec:component}
\subsection{Comparing Fixed Weighting Schemes With the Adaptive Weight}

In this section, we compare the adaptive weight ($\alpha=\frac{\text{avgConf}(\mathcal{D}_{test})}{ \text{avgConf}(\mathcal{D}_{o})}$) with setting $\alpha$ to fixed values $0.2$, $0.5$, and $0.9$. 
The difficulty level for the out-of-distribution situation is $10$. We evaluate the three calibration baselines on the six out-of-distribution test sets and one in-distribution test set using ResNet-$152$ as backbone and use the mean ECE value over all the seven test sets (six OOD datasets and one ID test set) as evaluation metric.

Fig. \ref{fig1} (left) indicates that for the Spline method, setting $\alpha$ to $0.2$ and $0.5$ deteriorates the Spline baseline, while $\alpha=0.9$ improves it. Because they are agnostic about test sets, setting fixed values might work in some proper cases and be less useful in others. Our design ($\alpha$) is shown to improve the baselines on various OOD test sets and seems to be superior to fixed values (under Spline). 

In Table \ref{alpha}, we also report the value of the adaptive $\alpha$ used in Table 1 and Table 2. It turns out that when the test set has low difficulty (ImageNet-V2-C) its distribution is closed to the distribution of training data, the computed $\alpha$ is closed to $1.0$, thus the in-distribution calibration performance is not compromised.

\subsection{Test Data Are Given in Batch}
In real-world scenarios, test data may not all be accessible. Here we study how the calibration performance changes when test data are given in batches of various sizes. In Fig. \ref{fig1} (right), $\alpha$ is calculated from test batches of various sizes. We observe our method still achieves improvement over the temperature scaling baseline and has similar ECE with the method computed on the full test set under reasonably large batch sizes ($\geq 64$).

\subsection{An Alternative Method}
\label{altern}
In L$210$-$216$ of the main paper, we mentioned that a possible way to calibrate OOD data is to estimate its difficulty and create a calibration set that has a closer difficulty level with the OOD test dataset.
Moreover, according to Sec. 3.5 of the main paper, the average confidence score could serve as an unsupervised indicator to the degree of how out-of-distribution a test set is \cite{guillory2021predicting}. 
Here, we propose another post-hoc calibration method for OOD calibration.
Specifically, we first estimate the error rate of a test set \cite{2022Leveraging}:
\begin{equation}
    error_{\mathcal{D}_{test}} = (1 - \text{Acc}(\mathcal{D}_{o})) + (\text{avgConf}(\mathcal{D}_{o})-\text{avgConf}(\mathcal{D}_{test})).
\end{equation}
Thus, we can compute $d_{\mathcal{D}_{test}}$ as:
\begin{equation}
    d_{\mathcal{D}_{test}} = \frac{error_{\mathcal{D}_{test}}}{1-error_{\mathcal{D}_{test}}}.
\end{equation}
According to Table \ref{estimate}, our estimation method is also shown to be effective. Specifically, it has the second lowest ECE on CIFAR-$10$-C and is only $0.0034$ higher than SVI-AvUC on ImageNet-C.

\subsection{Easy calibration set and hard calibration set have the same number of samples for tuning the function}
The size of $\mathcal{D}_h$ in our submission is $5,884$. We randomly sample the easy calibration set $\mathcal{D}_o$ into the same size ($5,884$), the difficulty of which remains the same due to random sampling. We report performance calibration (ECE, $\%$) of Temperature Scaling and our improved version on all the seven test sets below. The ResNet-$152$ classifier is used.
The results in Table \ref{table:same_size} show that our method remains beneficial, i.e., achieving lower ECE when combined with Temperature Scaling, when the easy and the hard calibration sets have the same size.
The results show that our method remains beneficial, \emph{i.e.,} achieving lower ECE when combined with Temperature Scaling, when the easy and the hard calibration sets have the same size.

\subsection{The original calibration set is not easy}
In Sec. 3.5 of main paper, we mentioned that difficulty is a relative concept and depends on the classifier. Note that for a weaker classifier, a certain dataset will be harder. With this in mind, we experimented with two weaker classifiers, (i.e., harder $\mathcal{D}_o$) and observed that our method is still effective. Specifically, we adopt LCNet-$050$ and TinyNet-E, which have $60.094\%$ and $59.856\%$ top-$1$ accuracy, respectively on the ImageNet-Val dataset. We apply Temperature Scaling with the proposed method to the two classifiers and report calibration performance (ECE, $\%$) below.
These results in Table \ref{table:hard_do} show that our method consistently improves Temperature Scaling when the ``easy calibration set'' has high difficulty (\emph{i.e.,} is not easy).

\subsection{More types of OOD test sets}
We further provide the calibration results (ECE, $\%$) on another challenging and diverse dataset iWildCam-WILDS \cite{koh2021wilds} with the ResNet-$50$ classifier. iWildCam-WILDS is an animal species classification dataset, where the distribution shift arises due to changes in camera angle, lighting, and background.
Tabel \ref{table:wilds} shows that our method can also improve the calibration performance on iWildCam-WILDS, especially, improves temperature scaling by $0.9\%$ decrease in ECE on the OOD test set.

\subsection{Other combination scheme of $\alpha$}
In the experiment section, we show the effectiveness of the simple linear combination of these two extreme logits. We further test another combination scheme in this section. According to Table \ref{table:comb}, it decreases ECE (\%) of uncalibration but is slightly worse than current scheme on ImageNet-V2 and ImageNet-R.

\setlength\tabcolsep{1.5pt}
\begin{table}[h]
\begin{center}
\begin{small}
\caption{Following the protocol in Gong \etal \cite{gong2021confidence}, we evaluate proposed ACE under domain generalization setting. We use Spline-based ACE and report ECE ($25$ bins, $\%$) for top-1 predictions.}
\label{table:dg}
\begin{tabular}{lcccccc}
\toprule
& Uncal. & Gong \etal [6] & ACE (Spline) \\
\midrule
A$\rightarrow$C & 11.84 & 12.53 & 4.82\\
A$\rightarrow$P & 6.81 & 5.56 & 2.84\\
A$\rightarrow$R & 4.31 & 6.25 & 3.77\\
\bottomrule
\end{tabular}
\end{small}
\end{center}
\end{table}

\subsection{ACE under the domain generalization setting}
In L$497$-L$503$, we discussed that we may have access to calibration datasets from multiple domains in realistic application scenarios. We here evaluate our ACE method with Spline baseline under domain generalization setting which has multiple source domains and compare with Gong \etal \cite{gong2021confidence}. Table \ref{table:dg} shows ACE achieves lower ECE compared with Gong \etal's method.

\end{document}